\documentclass{article}

\usepackage{microtype}

\usepackage{booktabs} 

\usepackage{hyperref}

\usepackage{float}
\usepackage{graphicx}
\usepackage{xcolor}
\usepackage{lipsum}
\usepackage{subcaption}

\usepackage{amsmath}
\usepackage{amssymb}
\usepackage{tikz}

\usepackage[accepted]{icml2020}

\newcounter{joanCounter}

\newcounter{cinjonCounter}

\newcounter{cosmasCounter}

\definecolor{Dgreen}{rgb}{0,0.6,0}
\newcounter{ronCounter}

\icmltitlerunning{In-Distribution Interpretability for Challenging Modalities}

\begin{document}
\setlength{\belowcaptionskip}{-10pt}

\twocolumn[
\icmltitle{In-Distribution Interpretability for Challenging Modalities}




\begin{icmlauthorlist}
\icmlauthor{Cosmas Hei{\ss}}{tub}
\icmlauthor{Ron Levie}{tub}
\icmlauthor{Cinjon Resnick}{nyu}
\icmlauthor{Gitta Kutyniok}{tub,tr}
\icmlauthor{Joan Bruna}{nyu,ias}

\end{icmlauthorlist}

\icmlaffiliation{tub}{Technical University of Berlin}

\icmlaffiliation{nyu}{New York University}

\icmlaffiliation{ias}{Institute for Advanced Study (IAS)}

\icmlaffiliation{tr}{University of Troms\o{}}

\icmlcorrespondingauthor{Cosmas Hei{\ss}}{heiss@math.tu-berlin.de}

\icmlkeywords{Machine Learning, ICML}

\vskip 0.3in
]



\printAffiliationsAndNotice{}  
\begin{abstract}
It is widely recognized that the predictions of deep neural networks are difficult to parse relative to simpler approaches. However, the development of methods to investigate the mode of operation of such models has advanced rapidly in the past few years. Recent work introduced an intuitive framework which utilizes generative models to improve on the meaningfulness of such explanations. In this work, we display the flexibility of this method to interpret diverse and challenging modalities: music and physical simulations of urban environments.
\end{abstract}
\section{Introduction}
While machine learning methods like linear regression or random forests allow for clear human interpretations, the explanations of why modern deep neural networks arrive at their decisions are much more opaque. However, the machinery required to explain how they reach their results has advanced a long way in the past five years \cite{Simonyan14a, NIPS2017_7062, DBLP:journals/corr/ShrikumarGK17, DBLP:journals/corr/RibeiroSG16, 10.1371/journal.pone.0130140}. The ideal is an approach that is domain-agnostic, method-agnostic, computationally efficient, and well-founded. We see these constraints satisfied in two methods  \citet{DBLP:journals/corr/abs-1905-11092,chang2018explaining} upon which we build. The underlying idea is to see which parts of the input must be retained in order to preserve the model's prediction, however in practice it is non-trivial to make sure the model's prediction remains meaningful. \citet{chang2018explaining} addresses this by using Generative Adversarial Networks (GANs, \cite{Goodfellow2014GenerativeAN,Pajot2019UnsupervisedAI,DBLP:journals/corr/abs-1801-07892,10.1145/3072959.3073659}) to infill images with feasible content. While in \cite{chang2018explaining} this method is used solely for image classification, we showcase the flexibility of such an approach.

\section{Method}
\label{sec:method}

Assume we are given access to a black box model $\Phi$ and a target datum $x$ for which we would like to understand why the model considers $\Phi(x) = \hat{y}$. We alter some aspect of $x$ with a transformation $T(x) = z$ in order to evaluate whether that aspect was important to the model's decision. This transformation $T$ is typically done by setting regions in the image to a background color \cite{DBLP:journals/corr/RibeiroSG16}, randomly flipping pixels \cite{DBLP:journals/corr/abs-1905-11092}, or by blurring specific regions \cite{DBLP:journals/corr/FongV17}.

In (\cite{DBLP:journals/corr/abs-1905-11092}, \cite{chang2018explaining}, \cite{DBLP:journals/corr/FongV17}), the authors evaluate whether\\$\Phi(x) = \Phi(z) = \hat{y}$ remains true after the transformation. \citet{DBLP:journals/corr/abs-1905-11092} connect this with Rate-Distortion theory and formulate the task of interpreting the model as finding a partition of components of $x$ into relevant and irrelevant components $S$ and $S^c$. For a probability distribution $\Upsilon$ defined on $[0, 1]^d$, from which random vectors $n \sim \Upsilon$ are drawn, the obfuscation $z$ of $x$ with respect to $S$ and $\Upsilon$ is characterized by $z_S = x_S$ and $z_{S^c} = n_{S^c}$.
Labeling the resulting distribution as $\Upsilon_S$, we arrive at an expected distortion of $S$ with respect to $\Phi$, $x$, and $\Upsilon$ referred to as Rate-Distortion Explanation (RDE):
\begin{align*}
D(S, \Phi, x, \Upsilon) = \frac12 \mathbb{E}_{z \sim \Upsilon_{S}}[(\Phi(x) - \Phi(z))^2]
\end{align*}
Similarly to how rate-distortion is used to analyze lossy data compression, RDE scores the components of $S$ according to the expected deviation from the original datum's classification score. The smallest set $S$ that ensures a limited distortion $D(S)$ will contain the most relevant components.
Intuitively, this means finding a minimal set of components in $x$ which cannot be randomly changed without affecting $\Phi$'s classification.
However, there is a subtle issue that reduces the effectiveness of this procedure that also affects other approaches like LIME \cite{DBLP:journals/corr/RibeiroSG16}.

The issue is that when we modify the image with random noise or setting parts of the image to a background color, we do not know if the obfuscation is still in the domain of $\Phi$. In other words, is $z$ close enough to the training data distribution for the minimization of $D$ to give meaningful information about why $\Phi$ made its decisions? If it is not close enough, then it is difficult to say that the classifier made its decision because that particular set of components in $S$ were important or not. It may instead be because $z$ is in a region of space for which the model has never developed a sufficient decision boundary.

\citet{chang2018explaining} addresses this by training an inpainting network $G$ to keep the range of $T$ as close as possible to the training data distribution $\mathcal{D}$. $G$ generates images such that a similarly powerful critic has trouble distinguishing whether the obfuscation $z(x, s, n) := T(x, s, n)$ (hereafter just $z$), which we can write as a convex combination using a binary mask $s$ as
\begin{align*}
    z := T(x, s, n) = x \odot s + G(x, s, n) \odot (1-s),
\end{align*}
came from $\mathcal{D}$.
Because $z$ equals $x$ where $x$ is masked by $s$, this can be viewed as sampling from the conditional data distribution $\mathcal{D}|_{z_s=x_s}$. While there are circumstances where this access to $\mathcal{D}$ is not possible, such as when examining a model packaged as part of a larger software installation, we consider it to be a reasonable request in most scenarios.


With that preface, we formalize our objective as follows. For $s \in \{0, 1\}^d$ and $z = T(x, s, n)$, we seek to minimize
\begin{align*}
L(s) = \frac12 \mathbb{E}_{z \sim \Upsilon_{s}}[(\Phi_d(x) - \Phi_d(z))^2] + \lambda ||s||_0.
\end{align*}
Although this reflects the goal of our optimization, obtaining a sparse mask $s$ and ensuring a low distortion, it is difficult to optimize due to the $\ell_0$-semi-norm. We therefore instead consider the following relaxation:
\begin{align}
\label{eq:loss}
L'(s) = \frac12 \mathbb{E}_{z \sim \Upsilon_{s}}[(\Phi_d(x) - \Phi_d(z))^2] + \lambda ||s||_1.
\end{align}

This is still difficult to optimize because $s$ is a hard binary mask representing the partition of $x$ into $x_S$ and $x_{S^c}$. Consequently, we relax further by following \cite{DBLP:journals/corr/abs-1905-11092} and formalizing the optimization approach as $s \in [0, 1]$ with the $\ell_1$-regularization encouraging saturation at either $0$ or $1$. We can then use SGD to optimize $L'$. This does not prevent $s$ from attaining non-extremal values, however we do not see that in practice.

Another way to circumvent this problem, presented in \cite{chang2018explaining}, is by viewing $s \sim \text{Bernoulli}(\theta)$. This can be done using the concrete distribution \cite{DBLP:journals/corr/MaddisonMT16,jang2016categorical}, which samples $s$ from a continuous relaxation of the Bernoulli distribution using some temperature $t$. We can then optimize the term \eqref{eq:loss} with respect to $\theta$ using SGD. Note that we still use $\ell_1$-regularization over $s\sim\text{Bernoulli}(\theta)$, which pushes the model to optimize for a sparse $s$.

Finally, a third way to try to get a sparse $s$ while minimizing the distortion is through Matching Pursuit (MP) \cite{MP93}. Here, components of $s$ are chosen in a greedy fashion according to which minimize the distortion the most. While this means that we have to test every component of $s$, it is applicable in cases where $s$ is low dimensional and we are only interested in few non-zero components. This is the case in our Radio Map experiments presented below.

\section{Experiments}

With our experiments, we demonstrate how capable this interpretability technique is for analyzing different data modalities. Whereas most works focus on images, we choose two challenging modalities that are often unexplored. The first is audio, where we focus on classification of acoustic instruments in the NSynth dataset \cite{engel2017neural}. In this setting, we train $G$ as described in Sec~\ref{sec:method} in order to inform $T$.

The second is interpreting the outcome of physical simulations used to estimate radio maps in urban environments. In this setting, we take a different tack with our inpainter. Because the data is expensive to gather, highly structured, and has capable associated physical simulations, we rely on a model-based approach along with heuristics to in-paint. We optimize $s$ with MP as described in Sec~\ref{sec:method}. 

\subsection{Audio}

We set $\mathcal{D}$ as the NSynth dataset \cite{engel2017neural}, a library of short audio samples of distinct notes played on a variety of instruments. The model $\Phi$ classifies acoustic instruments from $\mathcal{D}$. We note at this point that we follow the experimental setup of \cite{DBLP:journals/corr/abs-1905-11092} and compute the distortion with respect to the pre-softmax scores for each class. To train the inpainter $G$, we first sample $x$ from the dataset, $s$ as a random binary mask, and $n$ normally distributed to seed the generator. We then generate $x' = G(x, s, n)$ and have the discriminator adversarially differentiate whether $x'$ is real or generated.

We pre-process the data by computing the power-normalized magnitude spectrum and phase information using the DFT on a logarithmic scale from $20$ to $8000$ Hertz. We then train $G$ for $200$ epochs as a residual CNN with added noise in the input and deep features. While it did not fully converge, we found the outputs to be satisfactory, exemplified by the output in Fig~\ref{fig:gan_output}. More details regarding architecture are given in the appendix in Fig~\ref{fig:inpainter_diagram} and Table~\ref{tab:inpainter_layers}. 

\begin{figure}[ht]
    \centering
    \input{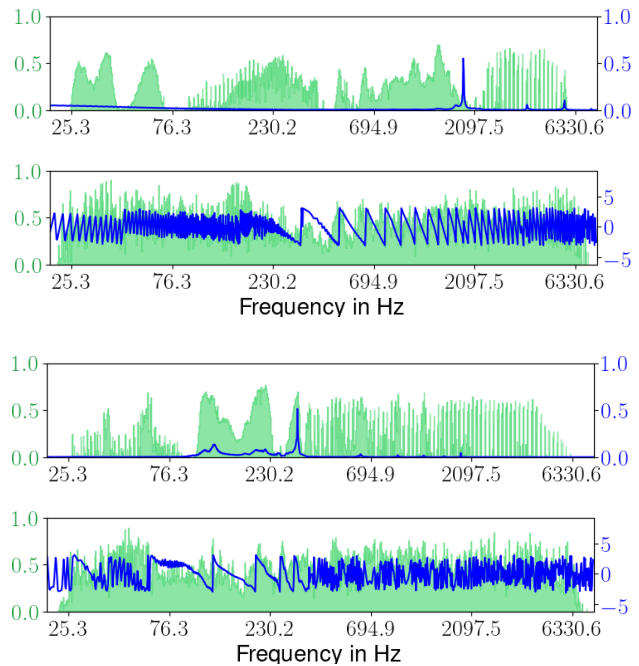}
    \caption{\textbf{Interpreting NSynth Model}: The optimized importance parameter $\theta$ (green) overlayed on top of the DFT (blue). For each of guitar and bass, the top graph shows the power-normalized magnitude and the bottom the phase. Notice the solid peaks between 30Hz and 60Hz for guitar and between 100Hz and 230Hz for bass. These are because the model is relying on those parts of the spectra, respectively, in order to classify as such. Notice also how many parts of the spectrum are important even when the magnitude is near zero. This indicates that the model pays attention to whether those frequencies are \emph{missing}.}
    \label{fig:nsynth_explainability}
\end{figure}

For computing the explainability maps, we constructed $\theta$ to be the Bernoulli variable dictating whether the phase or magnitude information of a certain frequency is dropped. $\theta$ was optimized to minimize Eq~\eqref{eq:loss} for $10^6$ iterations using the Adam optimizer with a step size of $10^{-5}$ and a regularization parameter of $\lambda = 50$. We used a temperature of $0.1$ for the concrete distribution. Two examples resulting from this process can be seen in Figure~\ref{fig:nsynth_explainability}.

Notice here that the method actually shows a strong reliance of the classifier on low frequencies (30Hz-60Hz) to classify the top sample in Figure~\ref{fig:nsynth_explainability} as a guitar, as only the guitar samples have this low frequency slope in the spectrum. We can also see in contrast that classifying the bass sample relies more on the continuous signal between 100Hz and 230Hz. Regarding the phase, it is interesting to see that if the phase angle changes smoothly with frequency, the model pays less attention than if the phase angle is changing rapidly. This can also be explained by only needing fewer samples throughout different frequencies to recognize smooth phase angle changes (versus rapid ones).

\begin{figure}[ht]
    \centering
    \input{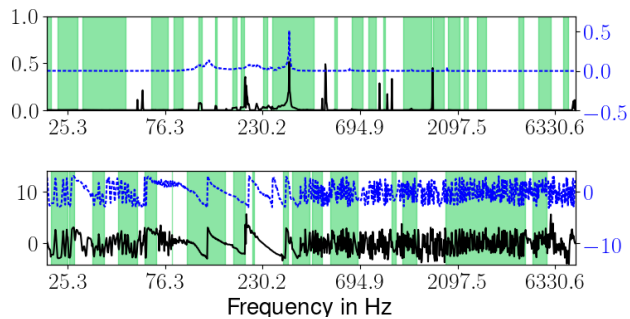}
    \caption{\textbf{Inpainted Bass}: Example inpainting from $G$. The random mask is zeroed out the green parts. The axes for the inpainted signal (black) and the original signal (blue dashed) are offset to improve visibility. Note how the inpainter generates plausible peaks in the magnitude and phase spectra, especially with regard to rapid ($\ge 600$Hz) vs. smooth ($< 270$Hz) changes in phase.}
    \label{fig:gan_output}
\end{figure}

\paragraph{Magnitude vs Phase}
We now consider the following interpretation query. By restricting $s$ to not act on each frequency but turn off or on the entire magnitude spectrum or phase information, we can probe the classifier for which of the two is more important. We can furthermore optimize the mask $s$ not only for one datum, but for all samples from a class. We can therefore extract the information if magnitude or phase is more important for predicting samples from a specific class.

For this, we again minimized \eqref{eq:loss} (meaned over all samples of a class) by optimizing $\theta$ as the Bernoulli parameter for $2 \times 10^5$ iterations using the Adam optimizer with a step size of $10^{-4}$ and the regularization parameter $\lambda=30$. Again, a temperature of $t=0.1$ was used for the concrete distribution. 

From the results of these computations, which can be seen in Table~\ref{tab:mag_vs_phase}, we can see that there is a clear difference on what the classifier bases its decision on across instruments. The classification of most instruments is largely based on phase information. For the mallet, the values are low for magnitude and phase, which means that the distortion is generally not really high compared to the regularization penalty even if the signal is completely inpainted. This underlines that due to the distortion being computed for pre-softmax scores, the regularization parameter generally has to be adjusted for every case.

\begin{table}[t]
\label{sample-table}
\vskip 0.15in
\begin{center}
\begin{footnotesize}
\begin{sc}
\begin{tabular}{lrr}
\toprule
Intrument & Magnitude & Phase \\
 &  Importance &  Importance\\
\midrule
Organ  & 0.829 & 1.0\\
Guitar & 0.0 & 0.999\\
Flute  & 0.092 & 1.0\\
Bass   & 1.0 & 1.0\\
Reed   & 0.136 & 1.0\\
Vocal  & 1.0 & 1.0\\
Mallet & 0.005   & 0.217\\
Brass  & 0.999 & 1.0\\
Keyboard & 0.003 & 1.0\\
String & 1.0 & 0.0\\
\bottomrule
\end{tabular}
\end{sc}
\end{footnotesize}
\end{center}
\caption{Magnitude Importance vs. Phase Importance.}
\label{tab:mag_vs_phase}
\vskip -0.1in
\end{table}

\subsection{Radio Maps}

In this setting, we assume a set of transmitting devices (TX) broadcasting a signal within a city. The received strength varies with location and depends on physical factors such as line of sight, reflection, and diffraction. The problem is to estimate the function that assigns the proper signal strength to each location in the city. Our dataset $\mathcal{D}$ is RadioMapSeer~\cite{levie2019radiounet} containing 700 maps, 80 TX per map, and a corresponding grayscale label encoding the signal strength at every location.

Our model $\Phi$ is a UNet~\cite{UNet15} architecture that receives as input three binary maps: a noisy map of the city where some buildings are missing, the TX locations, and some ground truth signal measurements. It is then trained to output the estimation of the signal strength throughout the city. We wish to understand whether signal measurements or buildings are more influential to our model's decisions.

\begin{figure}[h]
	\centering
	\begin{subfigure}{0.15\textwidth}\centering
		\includegraphics[width=0.997\linewidth]{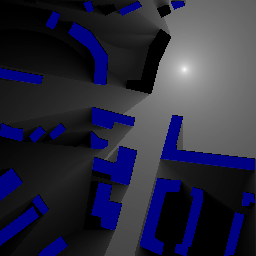}
		\caption{Ground Truth}
		\label{fig:radio1}
	\end{subfigure}%
	\hfill
	\begin{subfigure}{0.15\textwidth}\centering
		\includegraphics[width=0.997\linewidth]{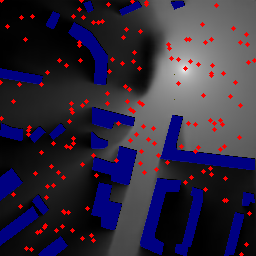}
		\caption{$\Phi$ Estimation}
		\label{fig:radio2}
	\end{subfigure}%
	\hfill
	\begin{subfigure}{0.15\textwidth}\centering
		\includegraphics[width=0.997\linewidth]{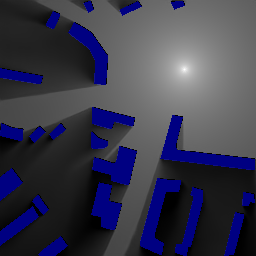}
		\caption{$\Phi_{\text{gt}}$ Estimation}
	   \label{fig:radio3}
	\end{subfigure}%
	\caption{\textbf{Radio map estimations}: The radio map (gray), input buildings (blue), and input measurements (red).}
	\label{fig:radio}
\end{figure}

We also consider a second model $\Phi_{\text{gt}}$ similar to the first except that it receives as input the ground truth city map along with the TX locations. Please see Fig~\ref{fig:radio1}, \ref{fig:radio2}, and \ref{fig:radio3} for examples of a ground truth map and estimations for $\Phi$ and $\Phi_{\text{gt}}$, respectively.


\paragraph{Explaining Radio Map $\Phi$}

\begin{figure}[t]
	\centering
	\begin{subfigure}{0.15\textwidth}\centering
		\includegraphics[width=0.997\linewidth]{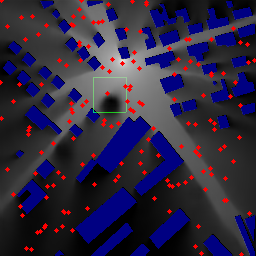}
		\caption{Estimated map. \newline \newline}
		\label{fig:radio_queries1}
	\end{subfigure}%
	\hfill
	\begin{subfigure}{0.15\textwidth}\centering
		\includegraphics[width=0.997\linewidth]{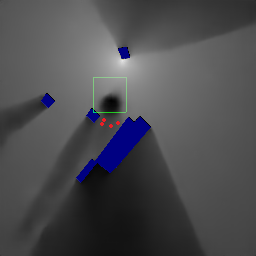}
		\caption{Explanation: Inpaint all unchosen measurements.}
		\label{fig:radio_queries2}
	\end{subfigure}%
	\hfill
	\begin{subfigure}{0.15\textwidth}\centering
		\includegraphics[width=0.997\linewidth]{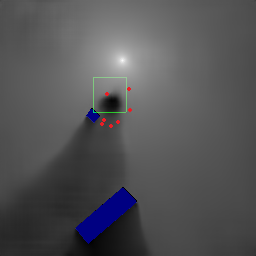}
		\caption{Explanation: Inpaint 2.5\% of unchosen measurements.}
	   \label{fig:radio_queries3}
	\end{subfigure}%
	\caption{\textbf{Radio map queries and explanations}: The radio map (gray), input buildings (blue), input measurements (red), and area of interest (green). Middle represents the query `How to fill in the image with shadows', while right is the query `How to fill in the image both with shadows and bright spot.' We inpaint with $\Phi_{\text{gt}}$.}
	\label{fig:radio_queries}
\end{figure}

Observe that in Fig~\ref{fig:radio1} and \ref{fig:radio2}, $\Phi$ interpolates the missing building with a shadow. As a black box method, it is unclear why it made this decision. Did it rely on signal measurements or on building patterns? To address this, we consider each building and measurement as potential targets for our mask $s$. 
As discussed in Sec~\ref{sec:method}, we use matching pursuit to find a minimal mask $s$ of decisive components (buildings or measurements). At each step, $s$ preserves what buildings and measurements it selects and zeroes out otherwise.

We consider two cases. The first is to accept the masked input from $s$ with the underlying assumption being that any subset of measurements and buildings is valid for a city map. 
For a fixed set of chosen buildings, adding more measurements to the mask typically brightens the resulting radio map. This lets us answer which measurements are most important for brightening the radio map.

On the other hand, we can inpaint with the trained $\Phi_{\text{gt}}$ to make a model-based prediction of the radio map conditioned on what $s$ preserved. This will overestimate the strength of the signal because there are fewer buildings to obstruct the transmissions. The more buildings that $s$ preserves, the less severe is the overestimate. We then sample this estimation to yield an in-painted map to input to $\Phi$. This lets us answer which measurements and buildings are most important to darkening the radio map. Between these two cases lay a continuum of completion methods where a random subset of the unchosen measurements are sampled from $\Phi_{\text{gt}}$ and the rest are set to zero.

Examples of these two cases are presented in Fig~\ref{fig:radio_queries} where we construct an explanation for a prediction $\hat{y} = \Phi(x)$. Note that we care about specific small patches exemplified by the green boxes. 

When the query is how to darken the free space signal (Fig~\ref{fig:radio_queries2}), the optimized mask $s$ suggests that samples in the shadow of the missing building are the most influential in the prediction. 
These dark measurements are supposed to be in line-of-sight of a TX, which indicates that the network deduced that there is a missing building. When the query is how to fill in the image both with shadows and bright spots (Fig~\ref{fig:radio_queries3}), both samples in the shadow of the missing building and samples right before the building are influential. This indicates that the network used the bright measurements in line-of-sight and avoided predicting an inordinately large building. To understand the chosen buildings, note that $\Phi$ is based on a composition of UNets and is thus interpreted as a procedure of extracting high level and global information from the inputs to synthesize the output. The locations of the chosen buildings in Figure \ref{fig:radio_queries} reflect this global nature.

\vspace{-2.5mm}
\section{Conclusion}
\vspace{-1.5mm}
In this paper, we have demonstrated that modern interpretability techniques can help explain a model's prediction in challenging domains like audio and physical simulations. This suggests using these techniques in areas across other modalities and especially within the experimental sciences, where interpretability is of utmost importance.


\vspace{-2.5mm}
\section{Acknowledgement}
\vspace{-1.5mm}
This work is partially supported by the Alfred P. Sloan Foundation, NSF RI-1816753, NSF CAREER CIF 1845360, NSF CHS-1901091, Samsung Electronics, the Institute for Advanced Study, and the Bundesministerium fur Bildung und Forschung (BMBF) through the Berlin Institute for the Foundations of Learning and Data (BIFOLD) and through Project MaGriDo. 

\clearpage

\bibliography{bibliography}
\bibliographystyle{icml2020}

\begin{appendix}
\begin{figure*}[h!]
    \centering
    \input{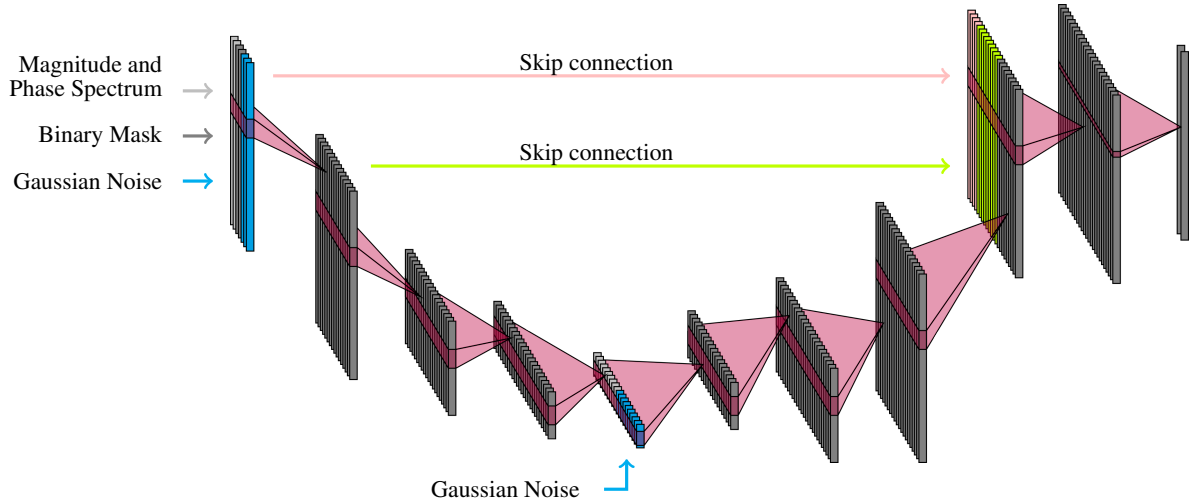}
    \caption{Diagram of the inpainting network for NSynth.}
    \label{fig:inpainter_diagram}
\end{figure*}
\begin{table*}[t]
\label{sample-table}
\vskip 0.15in
\begin{center}
\begin{small}
\begin{sc}
\begin{tabular}{lrrr}
\toprule
Layer & Filter Size & Output Shape & \# Params\\
\midrule
Conv1d-1        & 21  & [-1, 32, 1024] & 4,736\\
ReLU-2          &     & [-1, 32, 1024] & 0\\
Conv1d-3        & 21  & [-1, 64, 502]  & 43,072\\
ReLU-4          &     & [-1, 64, 502]  & 0\\
BatchNorm1d-5   &     & [-1, 64, 502]  & 128\\
Conv1d-6        & 21  & [-1, 128, 241] & 172,160\\
ReLU-7          &     & [-1, 128, 241] & 0\\
BatchNorm1d-8   &     & [-1, 128, 241] & 256\\
Conv1d-9        & 21  & [-1, 16, 112]  & 43,024\\
ReLU-10         &     & [-1, 16, 112]  & 0\\
BatchNorm1d-11  &     & [-1, 16, 112]  & 32\\
ConvTranspose1d-12& 21& [-1, 64, 243]  & 43,072\\
ReLU-13         &     & [-1, 64, 243]  & 0\\
BatchNorm1d-14  &     & [-1, 64, 243]  & 128\\
ConvTranspose1d-15& 21& [-1, 128, 505] & 172,160\\
ReLU-16         &     & [-1, 128, 505] & 0\\
BatchNorm1d-17  &     & [-1, 128, 505] & 256\\
ConvTranspose1d-18& 20& [-1, 64, 1024] & 163,904\\
ReLU-19         &     & [-1, 64, 1024] & 0\\
BatchNorm1d-20  &     & [-1, 64, 1024] & 128\\
Skip Connection &     & [-1, 103, 1024]& 0\\
Conv1d-21       & 7   & [-1, 128, 1024]& 92,416\\
ReLU-22         &     & [-1, 128, 1024]& 0 \\
Conv1d-23       & 7   & [-1, 2, 1024]  & 1,794\\
ReLU-24         &     & [-1, 2, 1024]  & 0\\
\bottomrule
Total number of parameters: & & & 737,266\\
\end{tabular}
\end{sc}
\end{small}
\end{center}
\caption{Layer table of the Inpainting model for the NSynth task.}
\label{tab:inpainter_layers}
\vskip -0.1in
\end{table*}
\end{appendix}

\end{document}